\documentclass[twocolumn,12pt]{asme2ej}

\usepackage{}

\usepackage{graphicx}
\usepackage{multirow}
\usepackage{lscape}
\usepackage{epsfig} 

\title{The NEOLIX Open Dataset for Autonomous Driving}

\author{
	\affiliation{Lichao Wang, Lanxin Lei, Hongli Song, Weibao Wang \\
		NEOLIX
	}	
}

\author{  
}

\begin{document}
	
	\maketitle    
	
	\begin{abstract}
		\centerline{Abstract}
		
		With the gradual maturity of 5G technology, autonomous driving technology has attracted more and more attention among the research community. Autonomous driving vehicles rely on the cooperation of artificial intelligence, visual computing, radar, monitoring equipment and GPS, which enables computers to operate motor vehicles automatically and safely without human interference.  However, the large-scale dataset for training and system evaluation is still a hot potato in the development of robust perception models. 
		In this paper, we present the NEOLIX dataset and its applications in the autonomous driving area. 
		Our dataset includes about 30,000 frames with point cloud labels, and more than 600k 3D bounding boxes with annotations. The data collection covers multiple regions, and various driving conditions, including day, night, dawn, dusk and sunny day.
		In order to label this complete dataset, we developed various tools and algorithms specified for each task to speed up the labelling process. It is expected that our dataset and related algorithms can support and motivate researchers for the further development of autonomous driving in the field of computer vision.

	\end{abstract}

	\section{Introduction}
	
	Autonomous driving technology can bring many benefits, such as better adapting to the crowd, improving road safety, easing traffic congestion and so on. Public large-scale datasets play an important role in accelerating the progress of computer perception tasks, including image classification, object detection, object tracking, semantic segmentation \cite{ref1,ref2,ref3} as well as lane detection.
	
	To further accelerate the development of autonomous driving technology, we present a diverse autonomous driving dataset, consisting of images recorded by 5 high-resolution cameras and point cloud data from 3 high-quality LiDAR scanners mounted on a fleet of self-driving vehicles. 
	
	The structure of this paper is organized as follows. Firstly, in Sec.2, we provide related work and have a comparison between our dataset and other published self-driving datasets. Then we elaborate sensor setup, design of LiDAR detection range and coordinate systems in Sec.3. In addition, we described the dataset from data collection, data format, data annotations and dataset analysis in Sec.4. What?s more, Sec.5 is our tasks \& metrics. Finally, we have our conclusion in Sec.6.

	\section{Related works}
	
	In recent years, autonomous driving datasets and their related algorithms have always been a hot topic. Works done by other companies on datasets and most related algorithms are summarized below.

	KITTI Dataset\cite{ref5,ref12}, which is jointly founded by Karlsruhr Institute of technology in Germany and Toyota American Institute of technology in 2012, is a classic computer vision algorithm evaluation dataset in the automatic driving scene in the world. The data set is used to evaluate the performance of stereo, optical flow, visual odometry, 3D object detection and 3D tracking in vehicle environment.
	
	SemanticKITTI Dataset\cite{ref6} is a supplement based on KITTI. It mainly provides a point annotation dataset of point cloud sequence. It has unprecedented class number and unknown detail level of each scan. The dataset contains 28 annotation categories, including cars, trucks, motorcycles, pedestrians and cyclists. In this way, the dynamic objects in the scene can be inferred.
	
	The ApolloScape Dataset\cite{ref8,ref11,ref14}, released in 2017, provides per-pixel semantic annotations for 140k camera images captured in various traffic conditions, ranging from simple scenes to more challenging scenes with many objects. The dataset further provides pose information with respect to static background point clouds.
	
	The Cityscapes Dataset\cite{ref7} is a new large-scale dataset contains a diverse set of stereo video sequences recorded in street scenes from 50 different cities. In addition to a larger set of 20000 weak annotation frames, it also has 5000 high-quality pixel-level annotation frames. As a result, this dataset is an order of magnitude larger than previous similar attempts.
	
	Waymo Open Dataset\cite{ref9} is a high-quality multi-mode annotation dataset for autonomous driving. It consists of height tagged data collected by Waymo self-driving vehicles and covers a wide range of environments, from dense urban centers to suburban landscapes.
	
	The Audi Autonomous Driving Dataset \cite{ref10}(A2D2) in 2020, features 2D semantic segmentation, 3D point clouds, 3D bounding boxes, and vehicle bus data. It includes more than 40,000 frames with semantic segmentation images and point cloud labels, more than 12,000 of them also have annotations for 3D bounding boxes.
	
	See Table~\ref{tab:my-table1} for a comparison of different datasets.
	
	\begin{table*}[]
		\centering
		\resizebox{\textwidth}{15mm}{
			\begin{tabular}{ccccccc}
				\hline
				& Annotation   LiDAR frames & 3D Boxes & 2D Boxes & LiDARs & Cameras & Average points/frame \\ \hline
				KITTI  & 15k                       & 80k      & 80k      & 1     & 4       & 120k                 \\ \hline
				Argo   & 22k                       & 993k     & -        & 2     & 9       & 107k                 \\ \hline
				Waymo  & 230k                      & 12M      & 9.9M     & 5     & 5       & 177k                 \\ \hline
				NEOLIX & 30k                       & 60k      & -        & 3     & 5       & 65k                  \\ \hline
		\end{tabular}}
		\caption{Comparison between our dataset and the other street-view self-driving datasets published}
		\label{tab:my-table1}
	\end{table*}

	\section{Sensor Setup}
	
	\subsection{Sensor Specifications}
	
	The data collection is conducted using three LiDAR sensors and five high-resolution cameras. Figure~\ref{fig_example1} is the layout of sensors. Detailed LiDAR data and camera image specifications are given below. 
	
	\begin{figure} 
		\centerline{\psfig{figure=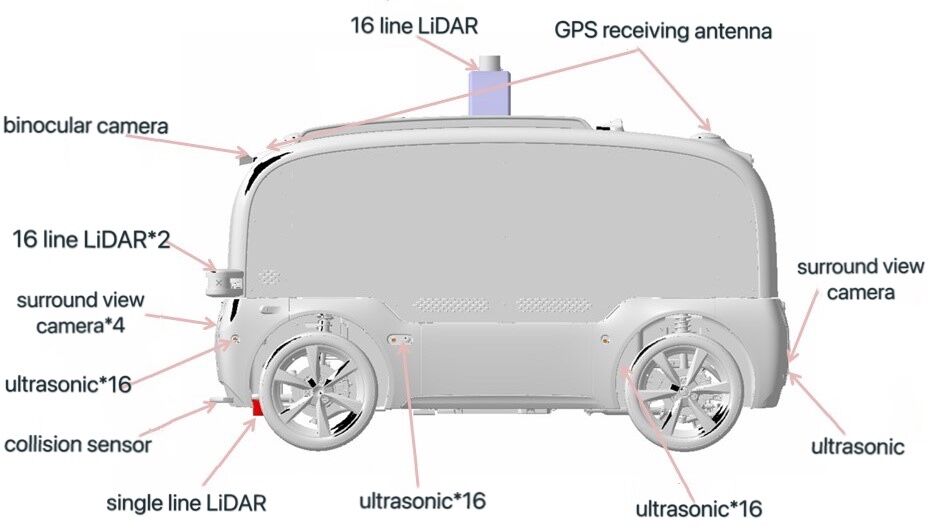,width=3.45in}}
		\caption{Sensor layout}
		\label{fig_example1}
	\end{figure}
	
	\subsection{Design of LiDAR detection range}
\begin{figure} 
		\centerline{\psfig{figure=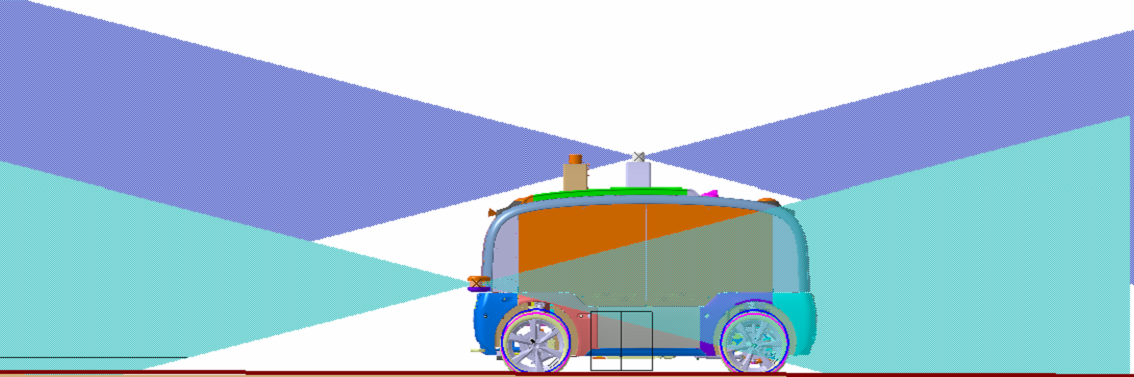,width=3.45in}}
		\caption{The side view of the LiDAR detection range}
		\label{fig_example2}
	\end{figure}

	There are three 16 line LiDARs, one on the roof, two of them on the front of the car. Figure~\ref{fig_example2} shows the side view of the LiDAR detection range. This kind of LiDAR arrangement can make the point cloud distribution denser and reduce the blind area.
	
	For the three LiDAR detection ranges, we test the vehicle with no load and full load,  respectively. Table~\ref{tab:my-table2} shows the size of the driverless vehicle, and Table~\ref{tab:my-table3} shows the detection range of the three LiDARs in the case of no load and full load of the vehicle.
	
	\begin{table}[]
		\centering
		\begin{tabular}{cc}
			\hline
			Parameter                 & Size(mm) \\ \hline
			The length of the vehicle & 2766.9   \\ \hline
			The width of the vehicle  & 1096.1   \\ \hline
			The height of the vehicle & 1824.6   \\ \hline
		\end{tabular}
		\caption{The size of the vehicle}
		\label{tab:my-table2}
	\end{table}

	\begin{table*}[]
		\centering
		\begin{tabular}{ccc}
			\hline
			Parameter                           & number(mm)    & number(mm)      \\ \hline
			16 line LiDAR height (vehicle roof) & 1821(no load) & 1790(full load) \\ \hline
			16 line LiDAR height (vehicle head) & 774(no load) & 741(full load)  \\ \hline
			Height of single line LiDAR         & 150(no load) & 106(full load) \\ \hline
		\end{tabular}
		\caption{The detection ranges of the three LiDARs in the case of no load and full load of the vehicle}
		\label{tab:my-table3}
	\end{table*}
	
	Sensor installation position information are show in Figure~\ref{fig_example3}, more details are given below.
	
	\begin{figure} 
		\centerline{\psfig{figure=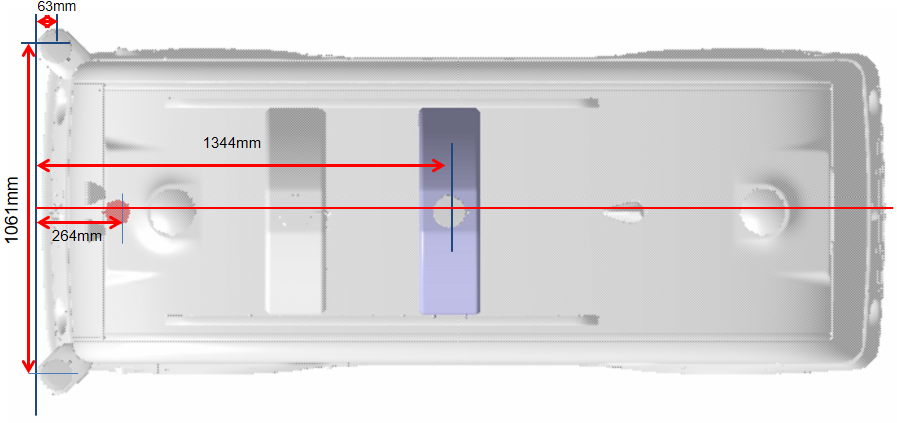,width=3.45in}}
		\caption{Sensor installation position information}
		\label{fig_example3}
	\end{figure}
	
	\begin{enumerate}
		\item The most forward part of the vehicle is the radar bracket installed at the front corner;
		\item The distance from the top radar center to the front end is 1343.7mm;
		\item The distance between the front corner mounted radar center and the front end is 63mm;
		\item The distance between the two front angle radars is 1061 mm;
		\item The distance from the center of the first line radar at the bottom to the front end is 264mm.
	\end{enumerate}

	\subsection{Coordinate Systems}
	This section describes the coordinate systems used in the dataset. All of the coordinate systems follow the right hand role.
	
	Geocentric inertial coordinate system:
	\begin{enumerate}
		\item The center of the earth inertial coordinate system is the origin at the center of the earth.
		\item The z-axis points to Polaris.
		\item The x-axis and y-axis are located in the equatorial plane, and the z-axis satisfies the right-hand rule and points to two stars.
		\item Inertial coordinate system often used as output of earth surface sensor.
	\end{enumerate}
	
	Vehicle coordinate system:
	\begin{enumerate}
		\item The origin is located at the mass center of the carrier and is fixed to the carrier (usually the center of the rear axle is selected).
		\item The x-axis points to the right along the axis of the carrier.
		\item The y-axis points forward.
		\item The z-axis points up.
	\end{enumerate}
	
	IMU coordinate system:
	\begin{enumerate}
		\item The origin is the coordinate origin of accelerometer and gyroscope.
		\item The x-axis, y-axis and z-axis are parallel to the corresponding axis of accelerometer and gyroscope.
		\item Since IMU is fixed with carrier, carrier coordinate is IMU coordinate system without considering installation error angle.
	\end{enumerate}
	
	Camera coordinate system:
	\begin{enumerate}
		\item The origin is the optical center of the camera.
		\item The x-axis and y-axis are parallel to the corresponding axis of the imaging plane coordinate system.
		\item The z-axis is the optical axis of the camera.
	\end{enumerate}
	
	LiDAR Spherical coordinate system:
	\begin{enumerate}
		\item The origin is located at the intersection of the rotation axes of the midpoint of the multi harness.
		\item The z-axis goes up along the axis.
		\item The x-axis and y-axis form a horizontal plane perpendicular to the z-axis.
	\end{enumerate}
	
	In our dataset, not only 3D detection but also other tasks to be completed, such as 2D detection and object tracking, are based on IMU coordinate system.


	\section{Dataset}
	
	The dataset annotates the data after fusion of three 16-line LiDAR point clouds. The data comes from Chaoyang Park, Waterfront park, scenes of operation in Xi'an and Shenzhen. We use three RoboSense LiDARs to label the ID ranged from 17859 to 26134, while the other IDs are corresponding to three Velodyne LiDARs.
	
	\subsection{Data collection}
	
	Among the four sites, we collected the maximum frames in Chaoyang Park, and the sampling period is to take one frame every 100ms. In Waterfront Park, we collected the minimum number of frames, and the sampling period is one frame every 500ms. In the operation scenes of Shenzhen and Xi'an, we used two different sampling periods. The specific collection duration and sampling frequency of the four data collection sites are shown in the Table~\ref{tab:my-table4}.
	
	\begin{table*}[]
		\centering
		\resizebox{\textwidth}{15mm}{
			\begin{tabular}{ccccc}
				\hline
				& frame number & the collection duration(min) & \multicolumn{2}{l}{the sampling frequency(frames / min)} \\ \hline
				Chaoyang Park            & 10000 & 20  & \multicolumn{2}{l}{600}           \\ \hline
				Waterfront park          & 3584  & 30  & \multicolumn{2}{l}{120}           \\ \hline
				Shenzhen operation scene & 8276  & 100 & 120(6736 frames) & 40(1540 frames) \\ \hline
				Xi'an operation scene    & 7859  & 80  & 120(6212 frames) & 60(1647 frames) \\ \hline
		\end{tabular}}
		\caption{Data collection}
		\label{tab:my-table4}
	\end{table*}

	\subsection{Data format}
	
	The dataset is divided into training set and testing set, and training set is divided into two parts. 	
	
	LiDAR: the point cloud data after fusion of three 16 line LIDARs are saved. After fusion and motion compensation, the three radars are finally saved in IMU coordinate system. In IMU coordinate system, the direction of car body forward is y-axis, clockwise rotation 90 degrees is x-axis, according to the right-hand rule, z-axis is vertical upward.
	
	Label: the result of obstacle labeling, which mainly stores the 3D information of obstacles.
	
	Compared with training set, the testing set has no label file and is mainly used to test the detection algorithm.
	
\begin{figure} 
		\centerline{\psfig{figure=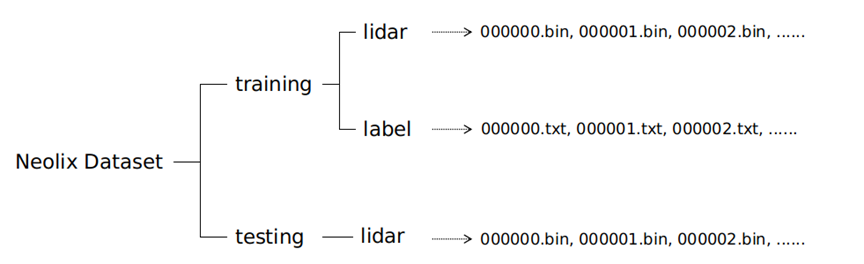,width=3.25in}}
		\caption{The overall frame structure of the dataset}
		\label{fig_example4}
	\end{figure}

	The overall frame structure of the dataset is shown in the Figure~\ref{fig_example4}.
	
	\subsection{Data Annotations}
	
	\begin{figure} 
		\centerline{\psfig{figure=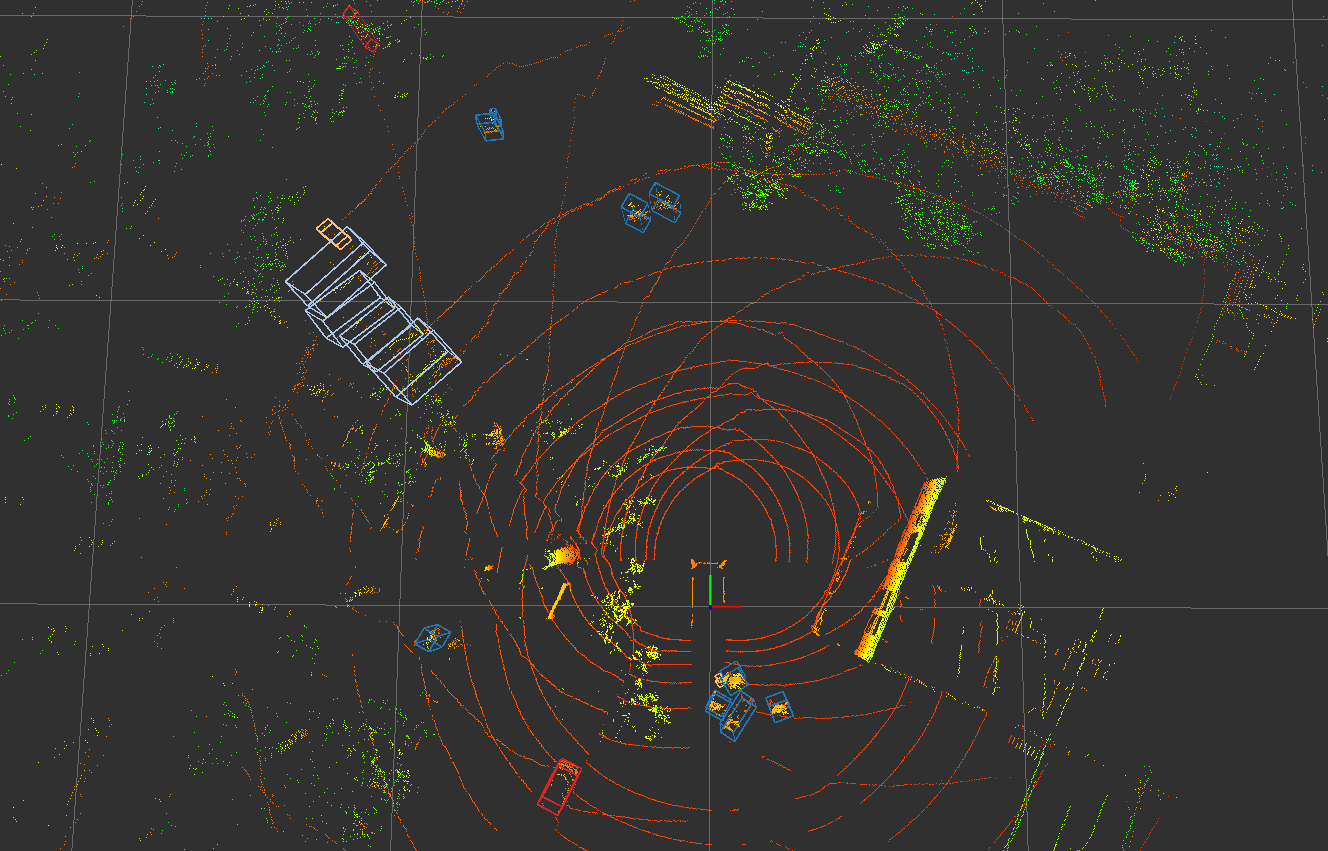,width=3.25in}}
		\caption{A frame point cloud and its corresponding bounding box}
		\label{fig_example5}
	\end{figure}

	We provide high-quality ground truth annotations, both for the LiDAR sensor readings as well as the camera images. For any label, we define length, width, height to be the sizes along x-axis, y-axis and z-axis, respectively.
	
	The dataset contains data from three LiDARs. We provide 3D bounding box labels in LiDAR data. The LiDAR labels are 3D 7-DOF (cx, cy, cz, l, w, h, $\theta$) bounding boxes in the vehicle frame with globally unique IDs, where cx, cy, cz represent the center coordinates, l, w, h are the length, width, height, and $\theta$ denotes the heading angle in radians of the bounding box. 
	
	Each frame of point cloud data in the label folder corresponds to a txt file. The annotation result of one frame of data is illustrated. Figure~\ref{fig_example5} shows a frame point cloud and its corresponding bounding box. Among them, different colors represent different categories of obstacles.

	Each label corresponding to a frame point cloud has several lines, and each line represents an obstacle. There are 15 fields in each line. The attributes of each field are as follows:
	
	
	0: Category. There are 15 categories in total 'Car', 'Bus', 'Truck', 'Trailer', 'Cyclist', 'Motorcyclist', 'Tricycle', 'Adult', 'Child', 'Animal', 'Barrier', 'Bicycle', 'Motorcycle', 'Bicycles', 'Unknown'.
	
	1: Truncation. It indicates whether the object is truncated by the edge of the image.
	
	2: Occlusion. It indicates the degree of occlusion of objects by other obstacles.
	
	3: Alpha. Angle of view.
	
	4 -- 7: The pixel coordinates of the top, bottom, left and right of the bounding box projected into the 2D image.
	
	8 -- 10: Characterizes the size of the 3D bounding box of the object. The order is height, width and length.
	
	11 -- 13: Coordinates of the center point of the bottom surface of the 3D bounding box. The order is x, y, z.
	
	14: Theta, which represents the angle from the head of the obstacle to the positive direction of x-axis.
	
	
	\begin{figure*} 
		\centerline{\psfig{figure=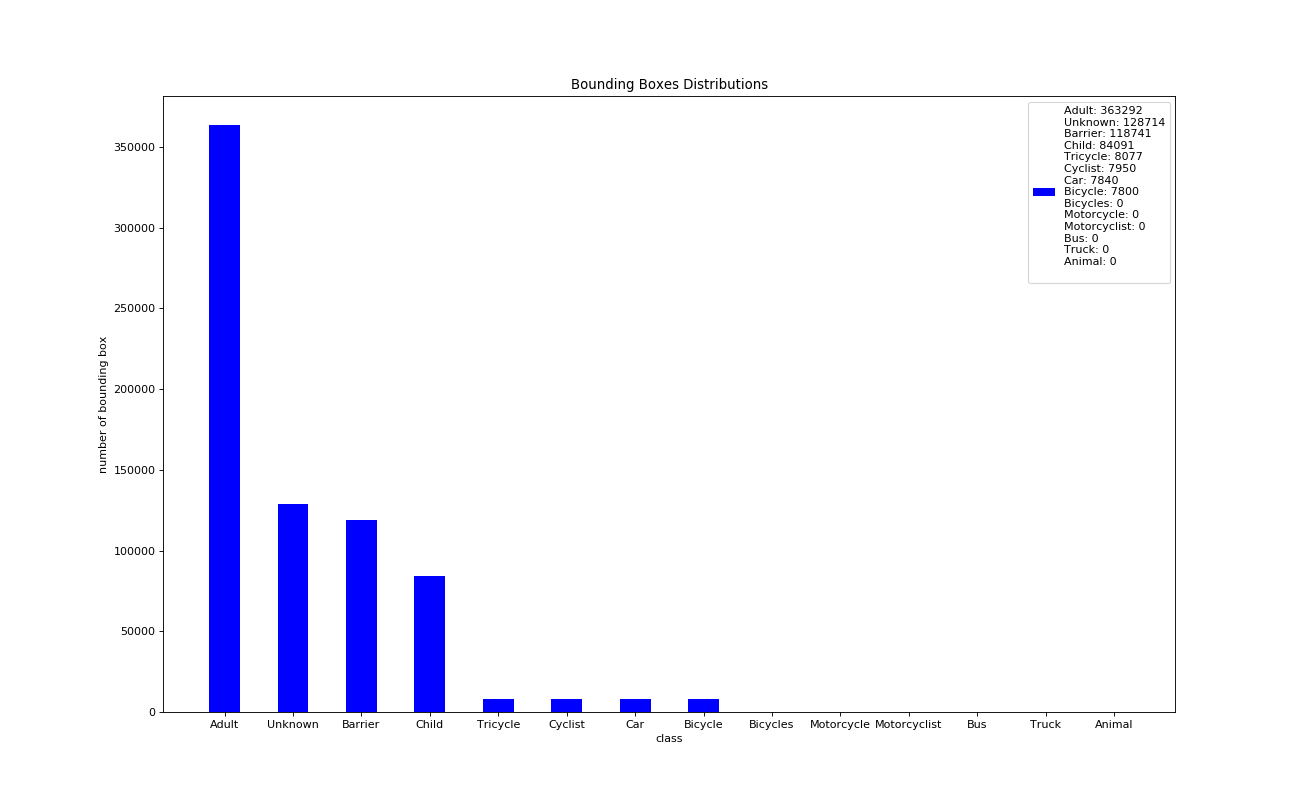,width=6.25in}}
		\caption{Bounding boxes distributions}
		\label{fig_example7}
	\end{figure*}
	
	\subsection{Dataset Analysis}
	
	Our dataset has scenes selected from different cities, including Beijing, Shenzhen and Xi?an, from different times of the day. There are more than 600k bounding boxes, more than 50\% of which are adults. See Figure~\ref{fig_example7} for the detailed bounding boxes distributions.

	\section{Tasks \& Metrics}
	
	\begin{table*}[]
		\centering
		\begin{tabular}{c|c}
			\hline
			Adult,Child                                              & Pedestrain \\ \hline
			Unknown,Barrier,Animal                                   & Unknown    \\ \hline
			Tricycle,Cyclist,Bicycle,Motorcycle,Bicycles,Motorcylist & Cyclist    \\ \hline
			Car,Bus,Truck,Trailer                                    & Vehicle    \\ \hline
		\end{tabular}
		\caption{The label categories classification}
		\label{tab:my-table6}
	\end{table*}

	\begin{table*}[]
		\centering
		\resizebox{\textwidth}{15mm}{
			\begin{tabular}{ccc}
				\hline
				& BEV(IOU threshold\_1/IOU   threshold\_2) & 3D((IOU threshold\_1/IOU   threshold\_2) \\ \hline
				pedestrian & 68.81/79.69 & 57.25/78.55 \\ \hline
				vehicle    & 49.48/58.77 & 28.97/50.10 \\ \hline
				cyclist    & 14.32/14.32 & 13.95/14.32 \\ \hline
				unknown    & 24.83/33.56 & 15.00/25.77 \\ \hline
		\end{tabular}}
		\caption{Baseline AP for different IOU thresholds.IOU threshold\_1 represents we use 0.5 IOU for pedestrian, 0.7 IOU for vehicle, 0.5 IOU for cyclist and 0.5 IOU for unknown while IOU threshold\_2 represents we use 0.25 IOU for pedestrian, 0.5 IOU for vehicle, 0.25 IOU for cyclist and 0.25 IOU for unknown.}
		\label{tab:my-table5}
	\end{table*}
	
	We define 3D object detection for the dataset and we also anticipate adding other tasks such as object tracking, semantic segmentation\cite{ref12,ref15}, behavior prediction, and imitative planning in the near future.

	\subsection{3D Detection}
	The 3D detection task involves predicting 3D upright boxes for vehicles, pedestrians, cyclists,  and unknown. AP(Average Precision), commonly used for object detection, does not have a notion of heading.

	\begin{equation}
	AP = 100\int_{0}^1 max\{p(r')|r'>=r\} dr
	\label{eq_1}
	\end{equation}

	where P represents precision, and R represents recall, p(r) is the P/R curve, which is a two-dimensional curve with precision and recall as vertical and horizontal coordinates.
	
	We provide baselines on our datasets based on recent approaches for detection, and in order to make the label classification clearer the label categories are classified as Table~\ref{tab:my-table6}. 
	
	The training set and validation set are divided according to the ratio of 1:9.Table~\ref{tab:my-table5} shows detailed results of the detection performance of each category on the validation set;

	For the BEV (Birds Eye View) and 3D metrics, we use two IOU thresholds to detect the performance of each category. We can clearly see that in both 3D method and BEV method, pedestrian has the highest AP value, while cyclist has the lowest AP value.

	\section{Conclusion}
	In this paper, we have presented a diverse dataset for autonomous driving by capturing a wide range of interesting scenarios, which shows the domain diversity among different cities, including Beijing, Shenzhen and Xi'an. The dataset and the corresponding code are open to the public; anyone interested in training is welcome.  We believe that this dataset will be highly useful in many areas of computer vision and autonomous driving.
	
	Our current dataset has yet to be developed in terms of data scale, tasks, acquisition equipment and so on. Here are some of our plans for the future: firstly, we plan to enlarge our dataset by collecting data under more diversified and complex driving environments including snow, and foggy. What?s more, we are also looking forward to completing tasks such as 2D detection, object tracking, and semantic segmentation\cite{ref12,ref15}. Last but not least, we will choose the most cost-effective acquisition equipment to improve the efficiency of data collection.


\end{document}